\documentclass[conference]{IEEEtran}
\IEEEoverridecommandlockouts
\usepackage{cite}
\usepackage{amsmath,amssymb,amsfonts}
\usepackage{algorithmic}
\usepackage{graphicx}
\usepackage{textcomp}
\usepackage{xcolor}
\usepackage{array}
\usepackage{multirow}
\usepackage{arydshln}
\newcolumntype{P}[1]{>{\centering\arraybackslash}p{#1}}
\usepackage{xcolor}
\usepackage{colortbl}
\usepackage[table]{xcolor}

\def\BibTeX{{\rm B\kern-.05em{\sc i\kern-.025em b}\kern-.08em
    T\kern-.1667em\lower.7ex\hbox{E}\kern-.125emX}}
\begin{document}

\title{TARAC: Mitigating Hallucination in LVLMs via \\ Temporal Attention Real-time Accumulative Connection}

% \author{Anonymous ICME submission}

\author{
\IEEEauthorblockN{Lei Jiang\textsuperscript{1} 
Chunzhao Xie\textsuperscript{1}
Tongxuan Liu\textsuperscript{1,3}
Yuting Zeng\textsuperscript{1,3}
Jinrong Guo\textsuperscript{3}
Yunheng Shen\textsuperscript{2,3}
Weizhe Huang\textsuperscript{1} \\
Jing Li\textsuperscript{1}
Xiaohua Xu\textsuperscript{1,*}}
\IEEEauthorblockA{\textsuperscript{1}University of Science and Technology of China \quad
\textsuperscript{2}Tsinghua University \quad
\textsuperscript{3}JD.com}
\thanks{*Corresponding author: Xiaohua Xu (xiaohuaxu@ustc.edu.cn). This work was supported in part by the National Natural Science Foundation of China (NSFC) with Grant No. 62172383 and No. 62231015, Anhui Provincial Key R\&D Program with Grant No. S202103a05020098, Research Launch Project of USTC with Grant No. KY0110000049.}
}

\maketitle

\begin{abstract}
Large Vision-Language Models (LVLMs) have demonstrated remarkable capabilities, yet they suffer from hallucinations that limit practical deployment. While various mitigation strategies exist, they often incur high computational overhead or require extensive retraining. In this paper, we address the issue of visual attention decay during generation, a key factor contributing to hallucinations. We propose Temporal Attention Real-time Accumulative Connection (TARAC), a novel training-free framework that dynamically accumulates and re-injects historical attention to sustain visual grounding. Inspired by cognitive reinforcement mechanisms, TARAC operates as a lightweight, plug-and-play module. Extensive experiments across diverse models (e.g., LLaVA, Qwen2-VL) and benchmarks demonstrate that TARAC significantly outperforms state-of-the-art methods. Remarkably, it achieves these gains with negligible inference overhead ($\sim$4\% TPOT increase), compared to the substantial costs of existing training-free baselines. Specifically, TARAC reduces hallucinated sentences by 25.2\% on CHAIR and improves Perception score by +10.65 on MME, validating its effectiveness and efficiency.
\end{abstract}


\section{Introduction}
In recent years, Large Vision-Language Models (LVLMs) \cite{liu2024visual,bai2023qwen,liu2024improved,chen2024far,wang2024qwen2,chen2024internvl} have demonstrated remarkable capabilities in vision-language understanding tasks and are rapidly evolving toward general-purpose vision-language models. However, studies such as \cite{liu2024survey,bai2024hallucination} have highlighted the significant challenge of hallucination in even the most advanced LVLMs, which substantially restricts their potential for practical applications.

To mitigate hallucinations, extensive research has been conducted. Training-based methods aim to enhance visual grounding by optimizing data quality (e.g., filtering detailed descriptions \cite{yue2024less}) or refining loss functions (e.g., reweighting generation loss \cite{cal}). Reinforcement learning approaches like LLaVA-RLHF \cite{sun2023aligning} and HA-DPO \cite{zhao2023beyond} have also been adapted to align model outputs with visual facts.

Training-free methods intervene during the inference stage. Some employ contrastive decoding strategies (e.g., VCD \cite{leng2024mitigating}, IBD \cite{zhu2024ibd}) to amplify visual reliance by contrasting logits from distorted inputs. Others, such as OPERA \cite{huang2024opera} and DOPRA \cite{wei2024dopra}, mitigate hallucinations by detecting and disrupting attention sinks on text tokens to prevent over-reliance on language priors. Additionally, agent-based frameworks like $V^*$ \cite{wu2024v} and LLaVA-PLUS \cite{liu2025llava} utilize iterative search or external tools to precisely locate image regions relevant to user queries, thereby improving response accuracy.

However, as noted in \cite{bai2024hallucination}, despite significant progress, hallucination remains a persistent challenge. Furthermore, the aforementioned training-free methods often rely on computationally expensive mechanisms like contrastive decoding or backtracking. These approaches inevitably introduce significant latency and computational overhead, limiting their practical deployment. To develop a more efficient solution, we investigate the underlying mechanism of hallucination generation.

\begin{figure}[t]
    \centering
    \includegraphics[width=0.48\textwidth]{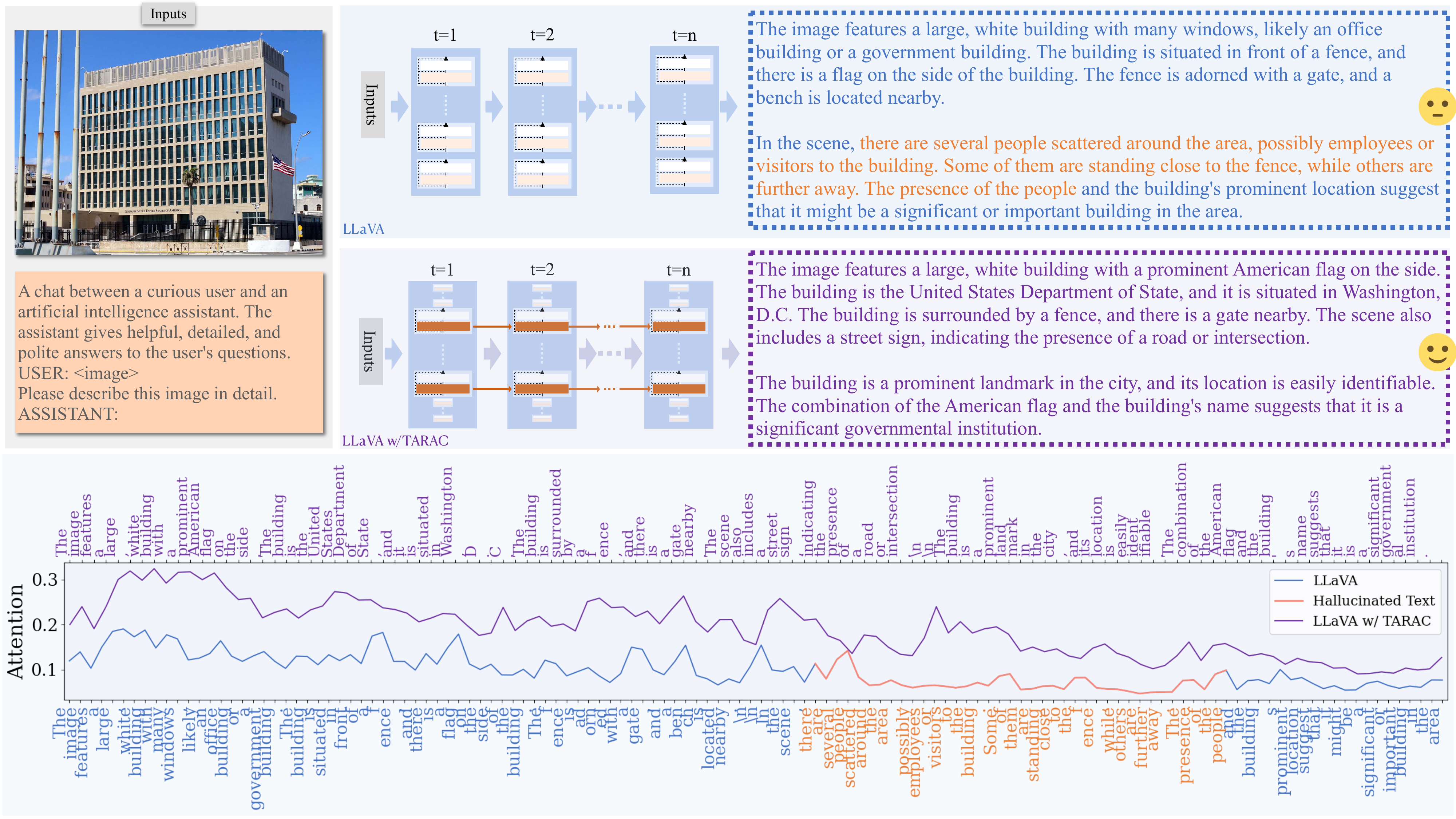}
    \vspace{-6mm}
    \caption{\textbf{Case Analysis.} LLaVA’s hallucinated responses and the correct responses with TARAC, along with the corresponding visual attention, are presented above. When visual attention is low, the model correspondingly generates hallucinated sentences. After applying TARAC, visual attention significantly increases, and the corresponding hallucinations no longer occur.}
    \label{fig:case}
    \vspace{-8mm}
\end{figure}

As shown in Figure \ref{fig:case}, we observe that during the generation process, LVLMs progressively allocate less attention to image tokens as more tokens are generated. Moreover, the positions where hallucinated statements occur exhibit correspondingly lower attention to visual information. Motivated by this observation, we propose a novel training-free approach, Temporal Attention Real-time Accumulative Connection (TARAC). Drawing inspiration from the forgetting curve in cognitive psychology \cite{ebbinghaus2013image}, where information retention naturally decays over time without reinforcement, TARAC addresses the analogous decay of visual attention during generation. Specifically, TARAC functions as a continuous reinforcement mechanism: it maintains a cumulative attention distribution and dynamically re-injects it to sustain the model's focus on visual content.

To summarize, the main contributions of this work are as follows:

\begin{enumerate}
\item We propose \textbf{TARAC}, a novel training-free framework designed to effectively mitigate hallucinations by counteracting visual attention decay. It dynamically accumulates and re-injects historical attention weights during generation to ensure sustained focus on visual content.

\item TARAC features \textbf{exceptional efficiency and broad applicability}. It introduces only $\sim$4\% TPOT (Time Per Output Token) increase and negligible memory overhead, significantly outperforming existing training-free methods that typically incur substantial computational overhead (e.g., $\sim$7.79$\times$ latency increase for backtracking-based approaches, or multiple forward passes for contrastive decoding methods). As a plug-and-play module, it consistently improves performance across diverse LVLM architectures (LLaVA-1.5, Qwen2-VL, InternVL2) without requiring architecture-specific modifications.

\item Comprehensive evaluations on multiple benchmarks demonstrate that TARAC achieves significant improvements: reducing hallucinated sentences by 25.2\% on CHAIR, improving Perception score by +10.65 on MME, establishing superior performance over existing approaches.
\end{enumerate}

\section{Related Work}
Existing approaches to mitigating hallucinations are categorized into training-based and training-free methods \cite{zhang2024seeing}.

\noindent{\textbf{Training-Based Method.}}
Training-based methods enhance visual grounding through data curation \cite{yue2024less}, specialized attention mechanisms \cite{zhao2024looking,xing2024mitigating}, contrastive learning \cite{cal}, and reinforcement learning techniques \cite{zhao2023beyond,sun2023aligning}. However, these methods demand substantial training resources and may not generalize well to unseen data.

\noindent{\textbf{Training-Free Method.}}
Training-free methods intervene during inference or employ agent-based strategies. Logit manipulation approaches \cite{leng2024mitigating,zhong2024investigating} and attention sink-based methods \cite{huang2024opera,wei2024dopra,zhang2024seeing} adjust generation probabilities or attention distributions to reduce hallucinations. Other interventions include manipulating attention heads \cite{yangmitigating}, stabilizing representations \cite{liu2024reducing}, or adjusting eigenspectrum variance \cite{tame}. Agent-based methods \cite{wu2024v,liu2025llava,liu2024chain,cao2024dualfocus} employ iterative processes or external tools to refine visual grounding. While effective, these methods often introduce significant inference overhead or rely on static interventions that do not account for the temporal dynamics of attention during generation.

Different from existing works, we identify that the decay of attention on image tokens during the generation process is a key factor contributing to hallucinations. To address this, we propose Temporal Attention Real-time Accumulation Connection (TARAC), a training-free framework that dynamically accumulates and reinforces attention on image tokens in real-time. Unlike static interventions, TARAC adapts to the temporal dynamics of generation, effectively maintaining visual grounding without the high cost of retraining or the latency of iterative agent-based pipelines.

\section{Methodology}
\begin{figure*}[!t]
    \centering
    \includegraphics[width=\linewidth]{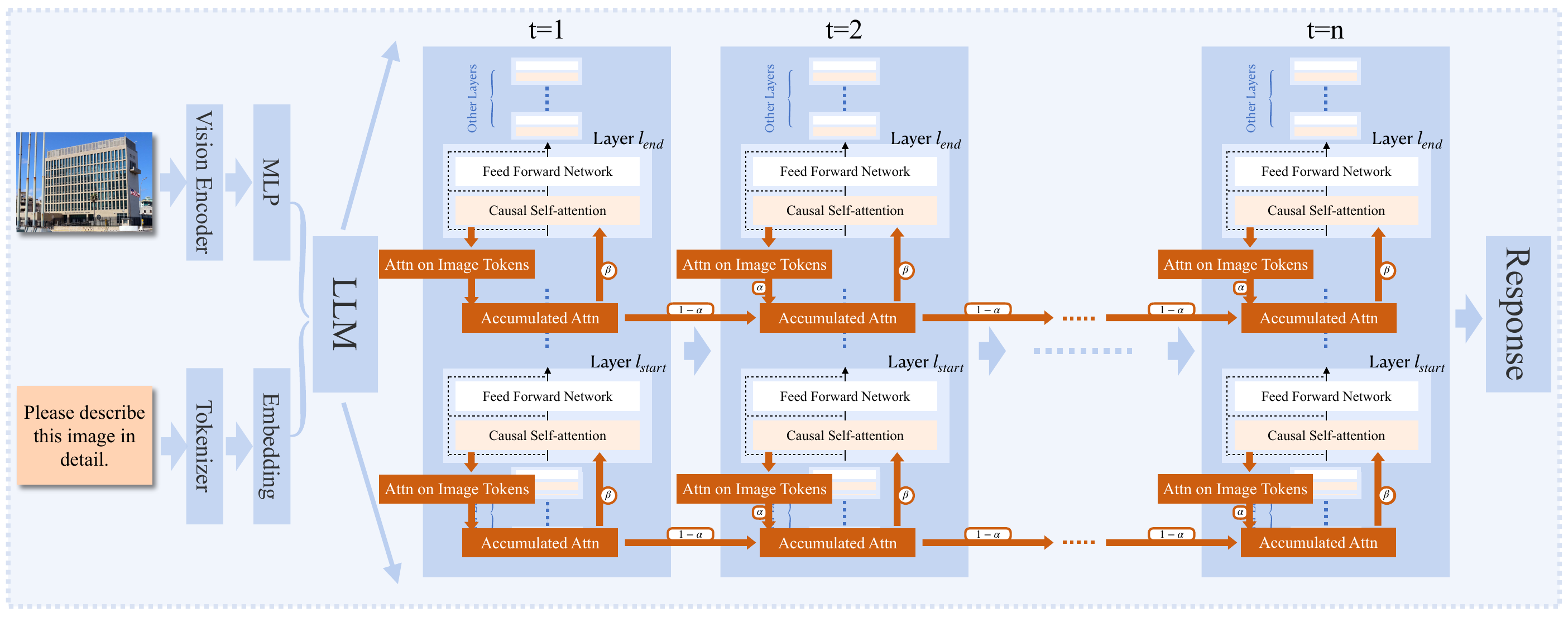}
    \vspace{-8mm}
    \caption{\textbf{Architecture of TARAC.} The framework operates within the Transformer's attention module (unfolded along the temporal dimension $t$). It enhances visual grounding by accumulating attention on image tokens and injecting this history into the current token's generation, followed by distribution renormalization.}
    \label{fig:structure}
    \vspace{-4mm}
\end{figure*}

Figure \ref{fig:structure} presents an overview of TARAC, which consists of three steps during the generation of each token. Firstly, TARAC accumulates the attention of the current generating token on the image tokens. Secondly, the accumulated attention on the image tokens is injected into the model’s inference process with a certain scaling factor. Thirdly, the attention distribution is renormalized to maintain the normalization property of the model’s attention.

\subsection{Accumulate Attention on Image Token} \label{method:record-attn}

For the Transformer's $l$-th layer with $H$ attention heads, the attention map generated during the inference process of the $t$-th token is denoted as $A_{l}^{t} \in \mathbb{R}^{H \times N_t \times N_t}$, where $N_t = N_i + N_p + t - 1$. Here, $N_p$ represents the number of tokens corresponding to the model's prompt, $N_i$ is the number of image tokens, and $t$ is the number of tokens generated by the model at time step $t$. The attention that needs to be recorded is the attention of current generating token at time step $t$ to the image tokens.
\begin{equation}
\bar{A}_{l}^{t} = \max_{j \in [1, 2, \cdots, H]} A_{l}^{t}[j, -1, i_s:i_e] \in \mathbb{R}^{1 \times 1 \times N_i}
\label{eq:captured-attn}
\end{equation}
 
Here, $j$ is the index of the attention head, and -1 indicates that we are only processing the current generating token. $i_s$ and $i_e$ represent the starting and ending indices of the image tokens, respectively. In this case, we compress the original attention across attention heads using the maximum value function. This approach extracts the prominent values of the original attention for each attention head, which can highlight the main visual information attended to by the model at that layer and shares this information across different heads. If the mean function were used instead, the prominent values might be averaged and overlooked, as they only appear in a few heads.

For the first generated token, we directly record its attention to the image tokens. For subsequent generated tokens, we update the accumulated attention using a memory update factor $\alpha \in [0,1]$. This both prevents the values of the accumulated attention from growing excessively and makes the model focus more on the attention of the most recently generated token to the image tokens. Here, $\alpha$ is similar to a mechanism that limits the size of the window for focusing on historical information, preventing excessively distant visual information from interfering with the current token generation. This decay mechanism draws inspiration from the Ebbinghaus Forgetting Curve \cite{ebbinghaus2013image}, which suggests that memory retention declines exponentially over time. By incorporating this, TARAC ensures that the model prioritizes relevant recent visual cues while gradually discarding outdated information. The formal representation of the accumulated attention $\hat{A}_{l}^{t} \in \mathbb{R}^{1 \times N_i}$ is as follows:
\begin{equation}
\hat{A}_{l}^{t} = 
\begin{cases} 
\bar{A}_{l}^{t}, & t = 1, \\
\alpha\bar{A}_{l}^{t} + (1 - \alpha) \hat{A}_{l}^{t-1}, & t > 1.
\end{cases}
\label{eq:accumulated-attn}
\end{equation}

\subsection{Inject Accumulated Attention} \label{method:inject-attn}
At time step $t$, the accumulated attention to image tokens given by Equation \ref{eq:accumulated-attn} is injected to the current generating token's attention to image tokens, enhancing the model's focus on visual information. This process can be formally expressed as follows:
\begin{equation}
A_{l}^{t}[:,-1,i_s:i_e] = A_{l}^t[:,-1,i_s:i_e] + \beta \cdot \hat{A}_{l}^{t}
\label{eq:injection}
\end{equation}

Where $\beta$ is the coefficient that controls the degree of accumulated attention injection. A larger $\beta$ may cause the model to pay more attention to visual information, but an excessively large value may lead to the injected accumulated attention dominating, resulting in repetitive generation. The mismatch in dimensions between $A_{l}^{t}[:,-1,i_s:i_e] \in \mathbb{R}^{H \times 1 \times N_i}$ and $\hat{A}_{l}^{t} \in \mathbb{R}^{1 \times 1 \times N_i}$ is due to the dimensionality reduction applied to the attention heads using the maximum function in Equation \ref{eq:captured-attn}. At this point, $\hat{A}_{l}^{t}$ is broadcast to each attention head in $A_{l}^{t}[:,-1,i_s:i_e]$.

\subsection{Re-normalizing Model's Attention}
Due to the modifications to the model's attention in Equation \ref{eq:injection}, in order to maintain the normalization property of the original attention matrix, we need to renormalize the last row of the attention matrix.  
Since the values in the original attention matrix become mostly close to zero after softmax normalization, applying softmax normalization again in this case will cause the last row's attention values for all tokens to approach $1/N_t$, thereby disrupting the model's original attention distribution.  
Therefore, we use row-sum normalization to enhance attention to image tokens while preserving the original attention distribution characteristics. This approach ensures the effective injection of accumulated attention while maximizing the retention of the model's original attention distribution. We formalize this process as follows:
\begin{equation}
A_{l}^t[j,-1,:] = \frac{A_{l}^t[j,-1,:]}{\sum_{m=1}^{N_t} A_{l}^t[j,-1,m]}, j=1,2,...,H
\end{equation}

In addition, since the model's ability to perceive visual information varies across different layers \cite{zhang2024cross,zhang2024redundancy}, our method is applied only within a specific layer range. The determination of this layer range is detailed in Section \ref{exp:ablation}.

\section{Experiments}
\setlength{\dashlinedash}{2pt}  % 线段长度
\setlength{\dashlinegap}{4pt}   % 线段间隔

\subsection{Experimental Setting}
\paragraph{Implementation Details.} We implement and test the TARAC method on LLaVA-1.5-7B \cite{liu2024visual}, Qwen2-VL-7B \cite{wang2024qwen2}, and InternVL2-8B \cite{chen2024internvl}.
In the experiment described in Section \ref{exp:main}, the TARAC applied layers are set to $l\texttt{=[10:16]}$ for LLaVA-1.5-7B, $l\texttt{=[14:22]}$ for Qwen2-VL-7B, and $l\texttt{=[8:16]}$ for InternVL2-8B.  
For the methods we compare, we use default parameters they
provided. Regarding the choice of hyperparameters for TARAC, we provide a detailed explanation using LLaVA as an example in Section \ref{exp:ablation}.

\paragraph{Benchmarks and Metrics.} CHAIR \cite{chair} is a metric for measuring Object Hallucination in image captioning tasks. CHAIR has two key dimensions: \textbf{$\mathrm{C}_I$}, which evaluates the proportion of hallucinated objects in the caption, and \textbf{$\mathrm{C}_S$}, which evaluates the proportion of hallucinated sentences in the caption. 
AMBER \cite{wang2023amber} is a benchmark for evaluating model hallucinations from both generative and discriminative task perspectives. 
SHR \cite{zhao2023beyond} uses a dataset of 200 images from the VG-100K dataset to evaluate image captions.
MME’s Coarse-Grained Subset assesses perception in existence, count, position, and color \cite{fu2023mme}. 

\subsection{Main Results} \label{exp:main}

We demonstrate TARAC's superiority through comprehensive comparisons with state-of-the-art methods across multiple dimensions: hallucination mitigation, inference efficiency, and text generation quality.

\noindent\textbf{Hallucination Mitigation Performance.}
We evaluate TARAC against existing methods on three generative benchmarks: CHAIR, AMBER, and SHR. As shown in Tables \ref{tab:chair-method}, \ref{tab:amber-mothod}, and \ref{tab:shr}, our method consistently outperforms VCD, AGLA, and OPERA across all metrics. On CHAIR, TARAC achieves substantial reductions of 25.2/16.6/17.2 in $C_S$ and 8.7/5.3/5.4 in $C_I$ compared to VCD/AGLA/OPERA respectively. On AMBER, despite a slight 1.2 decrease in \textit{Cover} (which we attribute to AMBER's more comprehensive dataset making it harder to rely on prior knowledge), TARAC significantly reduces hallucination rates. On SHR, our method surpasses VCD, AGLA, and OPERA by $\sim$31.3\%, $\sim$23.3\%, and $\sim$19.5\% respectively in hallucination sentence ratio, with GPT-eval confirming that the reduction aligns with human cognition. The consistent performance across these diverse benchmarks demonstrates that VCD, AGLA, and OPERA cannot continuously guide the model to focus on visual information in generative tasks, while TARAC effectively mitigates hallucination by enhancing visual attention and reducing reliance on language generation, though this leads to a controlled reduction in output length as it weakens the model's ``guesses" based on prior knowledge and co-occurrence reasoning \cite{leng2024mitigating}. The controlled reduction in generation length reflects TARAC's precision-oriented approach: by enhancing visual grounding, our method produces concise and accurate descriptions rather than verbose outputs that may contain hallucinated content. This "quality over quantity" trade-off is particularly valuable for applications requiring factual accuracy.

\begin{table}[h]
\centering
\vspace{-4mm}
\caption{CHAIR comparison on LLaVA-1.5. TARAC: $\alpha{=}0.3, \beta{=}0.9$.}
\label{tab:chair-method}
\vspace{-2mm}
\renewcommand{\arraystretch}{0.9}
\begin{tabular}{p{1cm}P{1.1cm}P{1.1cm}P{1.1cm}P{1.4cm}}
    \hline
    \small Method      & \small $C_S(\%)\downarrow$ & \small $C_I(\%)\downarrow$ & \small \small Recall(\%) & \small Avg. Len \\ \hline
    \small Greedy     & \underline{45.4} & \underline{13.4} & 77.5 & 89.09    \\
    \small Beam3      & 51.2 & 13.9 & 78.2 & 91.65    \\
    \small Beam5      & 49.6 & 13.8 & 76.7 & 93.01    \\ 
    \cdashline{1-5}
    \small DoLa(high) & 58.2 & 16.9 & \textbf{79.4} & 96.35    \\
    \small DoLa(low)  & 47.2 & 13.6 & 77.8 & 88.23    \\
    \small VCD        & 55.2 & 16.8 & 75.4 & 92.89    \\
    \small AGLA       & 46.6 & \underline{13.4} & \underline{78.5} & 91.69    \\
    \small OPERA      & 47.2 & 13.5 & 78.3 & 87.85    \\
    \rowcolor{gray!20}
    \small \textbf{Ours}       & \textbf{30.0} & \textbf{8.1}  & 72.0 & \textbf{83.24}    \\
    \hline
\end{tabular}
\vspace{-2mm}
\end{table}

\begin{table}[h]
\centering
\vspace{-4mm}
\caption{AMBER on LLaVA-1.5 (same hyperparameters as Table~\ref{tab:chair-method}).}
\label{tab:amber-mothod}
\vspace{-2mm}
\renewcommand{\arraystretch}{0.9}
\begin{tabular}{p{1cm}P{1.6cm}P{1.4cm}P{1cm}P{1cm}}
\hline
Method     &  \textbf{\footnotesize $\mathrm{CHAIR}(\%)\downarrow$} & \textbf{\footnotesize $\mathrm{Cover}(\%)\uparrow$} & \textbf{\footnotesize $\mathrm{Hal}(\%)\downarrow$}  & \textbf{\footnotesize $\mathrm{Cog}(\%)\downarrow$} \\ \hline
\small Greedy     & 7.6   & 49.5  & 32.1 & 3.8 \\
\small Beam3      & 7.9   & 49.7  & 37.5 & 4.6 \\
\small Beam5      & 8.9   & 48.8  & 38.1 & 4.8 \\
\cdashline{1-5}
\small DoLa(high) & 8.8   & \textbf{52.2}  & 40.0 & 4.2 \\
\small DoLa(low)  & 7.4   & 50.7  & 33.3 & 3.9 \\
\small VCD        & 8.7   & \underline{51.5}  & 41.1 & 4.4 \\
\small AGLA       & 7.4   & 51.1  & 34.6 & 3.9 \\
\small OPERA      & \underline{6.4}   & 49.7  & \underline{29.1} & \underline{2.9} \\
\rowcolor{gray!20}
\small \textbf{Ours}       & \textbf{5.0}   &  48.3  & \textbf{27.1} & \textbf{2.5} \\ \hline
\end{tabular}
\vspace{-2mm}
\end{table}

\begin{table}[h]
\centering
\caption{SHR on LLaVA-1.5. SPI/WPI: sentences/words per image; HSPI/HWPI: hallucinated sentences/words per image; HSR/HWR: hallucination sentence/word ratio.}
\label{tab:shr}
\vspace{-2mm}
\begin{tabular}{p{1.2cm}P{0.5cm}P{0.7cm}P{0.7cm}P{0.7cm}P{0.7cm}P{0.7cm}}
\hline
\small Method     & \small SPI↑ & \small WPI↑ & \small HSPI↓ & \small HWPI↓ & \small HSR↓ & \small HWR↓ \\ \hline
\small Greedy     & 5.08               & 89.73              & 2.1                      & 39.76                    & 0.42                & 0.45                \\
\small Beam3      & \underline{5.15}               & \underline{93.96}              & 2.2                      & 42.97                    & 0.43                & 0.46                \\
\small Beam5      & 5.12               & 93.81              & 2.26                     & 44.33                    & 0.44                & 0.47                \\
\cdashline{1-7}
\small DoLa(low)  & 5.02               & 89.92              & 2.04                     & 38.79                    & \underline{0.41}                & \underline{0.43}                \\
\small DoLa(high) & \textbf{5.36}               & \textbf{95.69}              & 2.36                     & 44.55                    & 0.44                & 0.47                \\
\small VCD        & 5.08               & 90.54              & 2.4                      & 45.44                    & 0.48                & 0.51                \\
\small AGLA       & 4.95               & 88.42              & 2.11                     & 39.96                    & 0.43                & 0.46                \\
\small OPERA      & 4.79               & 85.67              & \underline{1.95}                     & \underline{37.18}                    & \underline{0.41}                & 0.44                \\
\rowcolor{gray!20}
\small \textbf{Ours}      & 4.93               & \textbf{82.75}              & \textbf{1.64}                     & \textbf{28.57}                    & \textbf{0.33}                & \textbf{0.35}               \\ \hline
\end{tabular}
\vspace{-2mm}
\end{table}

\begin{figure}[h]
    \centering
    \vspace{-8mm}
    \includegraphics[width=\linewidth]{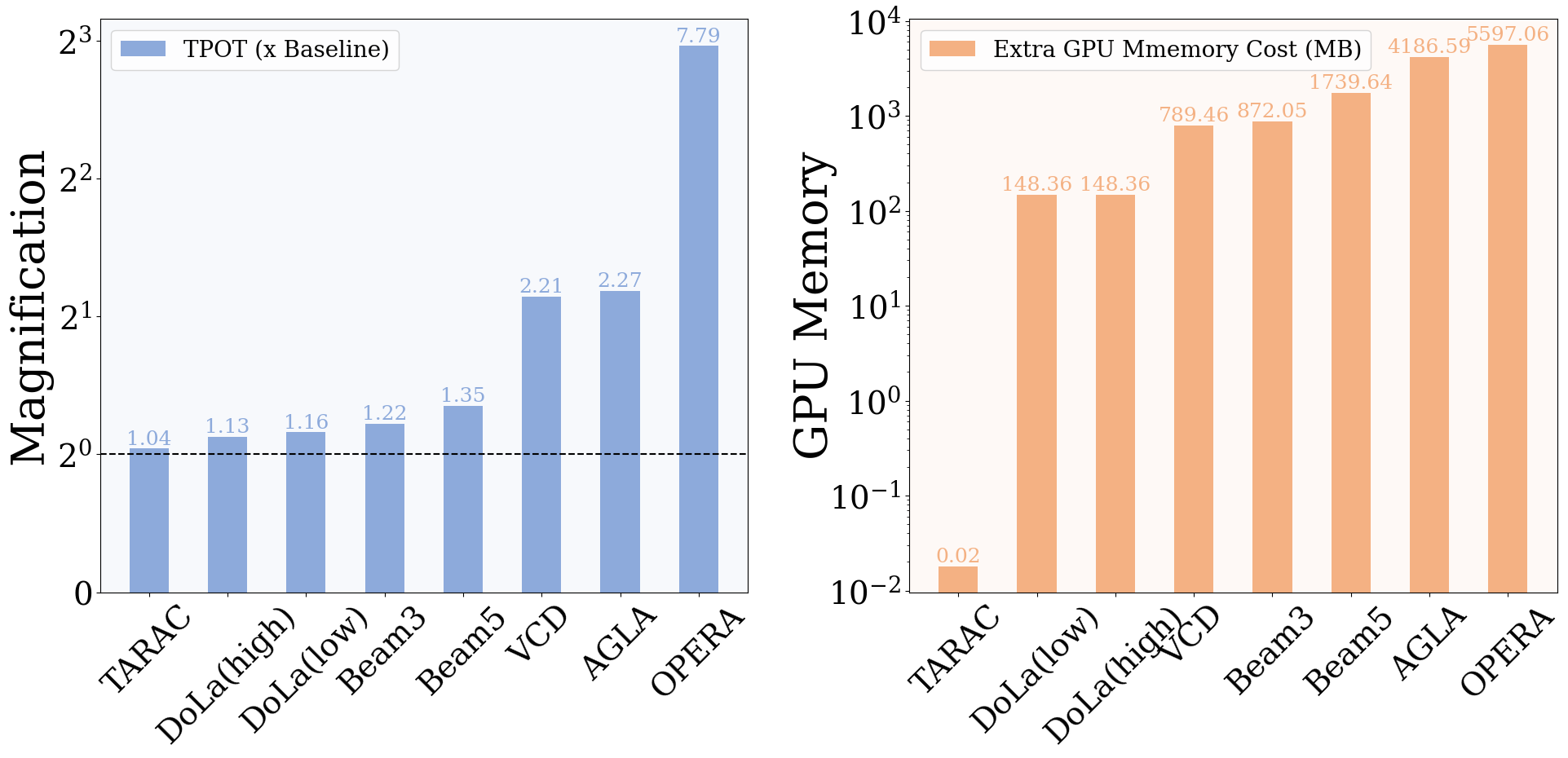}
    \vspace{-8mm}
    \caption{Inference efficiency: Time Per Output Token (TPOT, multiple of baseline) and GPU memory increase (MB).}
    \label{fig:cost}
\end{figure}

\noindent\textbf{Inference Efficiency.}
To evaluate the computational overhead of TARAC, we conduct efficiency analysis on LLaVA-1.5 using the prompt "Please describe the image in detail." Figure \ref{fig:cost} presents a comprehensive comparison of Time Per Output Token (TPOT) and GPU memory consumption across different methods. TARAC demonstrates remarkable efficiency, introducing only a minimal 4\% increase in inference time while maintaining virtually zero additional GPU memory overhead (0.02 MB). This efficiency advantage stems from TARAC's lightweight design: unlike contrastive decoding methods (VCD, AGLA) that require generating additional logits or backtracking mechanisms (OPERA), TARAC simply accumulates attention weights on image tokens during forward passes and injects this information to guide subsequent generation. In contrast, competing methods show substantially higher computational costs—OPERA incurs a 7.79× time overhead, while VCD and AGLA require 2.21× and 2.27× more time respectively, accompanied by significant memory increases ranging from 148 MB to over 5 GB.

\noindent\textbf{Text Generation Quality.}
To assess whether hallucination mitigation comes at the cost of text quality, we evaluate the linguistic fluency of generated captions using perplexity analysis. We compute perplexity scores for captions generated by different methods on the AMBER dataset, employing four GPT-2 variants of increasing complexity (base, medium, large, and XL) as evaluation models. Table \ref{tab:ppl} reveals that TARAC achieves superior text quality compared to all competing hallucination mitigation approaches. Specifically, TARAC consistently outperforms VCD, AGLA, DoLa, and OPERA across all model scales, with perplexity reductions ranging from 6.7\% to 22.5\% depending on the evaluator model size. Notably, TARAC even approaches the fluency levels of beam search methods (Beam3/Beam5), which prioritize text quality over hallucination control. This demonstrates that TARAC's attention-based approach not only reduces hallucinations but also preserves—and in many cases enhances—the natural flow and coherence of generated text, achieving an optimal balance between factual accuracy and linguistic quality.

\begin{table}[h]
\centering
\vspace{-4mm}
\caption{Perplexity comparison ($PPL_{1\text{-}4}$: GPT-2, -medium, -large, -xl).}
\label{tab:ppl}
\vspace{-2mm}
\begin{tabular}{lP{1cm}P{1cm}P{1cm}P{1cm}}
\hline
Method   & $PPL_1\downarrow$   & $PPL_2\downarrow$ & $PPL_3\downarrow$ & $PPL_4\downarrow$ \\ \hline
Greedy     & 14.01 & 11.23 & 10.13 & 9.46   \\
Beam3      & 12.95 & 10.29 & 9.30  & 8.73   \\
Beam5      & 12.50 & 9.95  & 9.02  & 8.46   \\
\cdashline{1-5}
DoLa(low)  & 14.26 & 11.46 & 10.33 & 9.64   \\
DoLa(high) & 17.44 & 13.82 & 12.41 & 11.57  \\
VCD        & 16.20 & 12.91 & 11.70 & 10.95  \\
AGLA       & 14.36 & 11.47 & 10.41 & 9.69   \\
OPERA      & 14.13 & 11.24 & 10.19 & 9.58   \\
\rowcolor{gray!20}
\textbf{Ours}      & 13.13 & 10.67 & 9.61  & 9.01  \\ \hline
\end{tabular}
\vspace{-4mm}
\end{table}

\subsection{Ablation Study and Analysis} \label{exp:ablation}
To provide deeper insights into TARAC's effectiveness and design choices, we conduct comprehensive ablation studies across four key dimensions. We first analyze the visual attention mechanism to understand how TARAC affects model behavior, then evaluate cross-model generalizability to validate broad applicability, investigate optimal layer selection for implementation guidance, and finally examine hyperparameter sensitivity for practical deployment. These analyses collectively demonstrate TARAC's robustness and provide clear guidance for its application across different LVLM architectures.

\begin{figure}[h]
\vspace{-4mm}
\includegraphics[width=\linewidth]{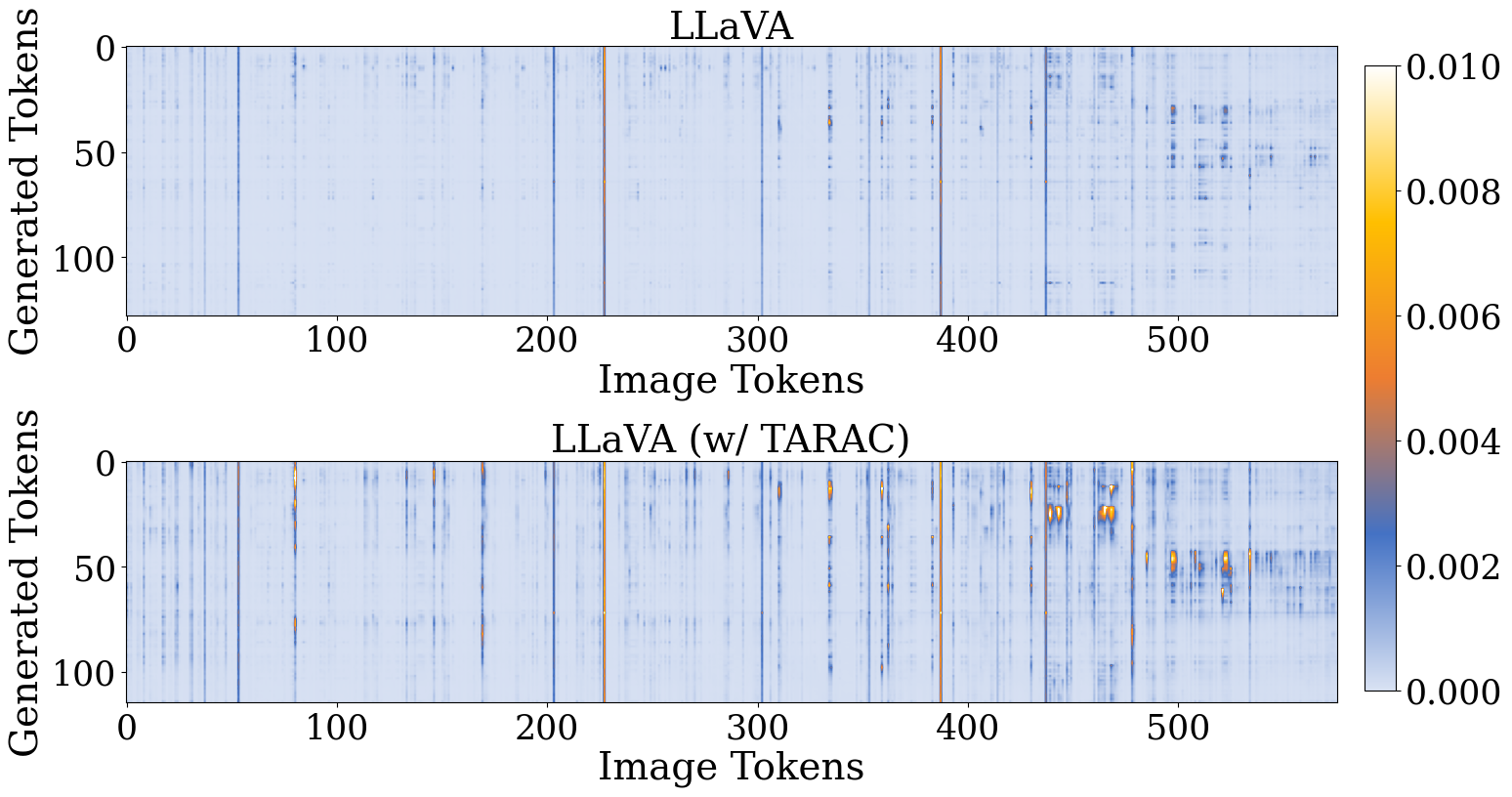}
    \vspace{-8mm}
    \caption{Image token attention w/ and w/o TARAC, enhanced attention sink.}
    \label{fig:sink}
\end{figure}

\noindent\textbf{Visual Attention Mechanism Analysis.}
Prior work \cite{zhang2024seeing} demonstrates that stronger attention sink on image tokens is negatively correlated with hallucination. 
Figure \ref{fig:sink} shows that TARAC substantially increases the attention sink on image tokens compared with vanilla LLaVA, exhibiting the same trend. 
Additional attention visualizations provided in the Appendix qualitatively support this observation. 
Taken together, these results suggest that TARAC mitigates hallucinations by reinforcing and stabilizing the model’s reliance on visual evidence during generation.    

\noindent\textbf{Cross-Model Generalizability.}
To validate the robustness of TARAC across different architectures and task types, we evaluate it on the MME benchmark using LLaVA-1.5, Qwen2-VL, and InternVL2.
As shown in Table \ref{tab:mme}, TARAC consistently improves the Perception scores across all models (e.g., +14.18 for LLaVA-1.5, +10.65 for Qwen2-VL), detailed breakdown in Appendix 
This indicates that enhancing visual attention effectively sharpens the model's discriminative capabilities.
More details on generative tasks such as CHAIR and AMBER can be found in the Appendix.
\begin{table}[htbp]
    \centering
    \vspace{-4mm}
    \caption{MME on different models w/ and w/o TARAC ($\alpha{=}0.5, \beta{=}0.5$).}
    \label{tab:mme}
    \vspace{-2mm}
    \renewcommand{\arraystretch}{1.1}
    \begin{tabular}{llcc}
    \hline
    \small Model & \small Method & \small Cognition$\uparrow$ & \small Perception$\uparrow$ \\ \hline
    \multirow{2}{*}{LLaVA-1.5} & Greedy & 323.57 & 1482.96 \\
                               & \textbf{Ours} & \textbf{339.29} & \textbf{1497.14} \\ \hline
    \multirow{2}{*}{Qwen2VL}   & Greedy & 556.43 & 1649.66 \\
                               & \textbf{Ours} & \textbf{581.07} & \textbf{1660.31} \\ \hline
    \multirow{2}{*}{InternVL2} & Greedy & 566.43 & 1599.84 \\
                               & \textbf{Ours} & \textbf{566.43} & \textbf{1615.69} \\ \hline
    \end{tabular}
    \vspace{-2mm}
    \end{table}

\noindent\textbf{Layer Selection Analysis.}
Using MME as the evaluation benchmark, Table \ref{tab:mme-layer} reveals that middle layers (10-16) achieve optimal performance with relatively balanced improvements in both \textit{Cognition} and \textit{Perception} scores. 
Applying TARAC to shallow layers (1-10) actually decreases \textit{Perception} scores by 16.13 points. This is likely because attention in shallow layers is typically distributed uniformly to gather global context \cite{zhang2024cross}, making early intervention unnecessary and potentially disruptive.

\vspace{-2mm}
\begin{table}[h]
\centering
\vspace{-4mm}
\caption{Impact of layer selection on MME scores (LLaVA-1.5). Detailed breakdown in Appendix.}
\label{tab:mme-layer}
\vspace{-2mm}
\renewcommand{\arraystretch}{1.1}
\begin{tabular}{lcc}
\hline
Layer Range & Cognition & Perception (Total) \\ \hline
baseline    & 323.57    & 1482.96            \\
{[}1:10{]}  & 307.86    & 1466.83            \\
{[}8:16{]}  & \underline{332.86}    & \underline{1499.89}            \\
{[}10:16{]} & \textbf{339.29}    & 1497.14            \\
{[}10:18{]} & 327.14    & 1494.27            \\
{[}10:20{]} & 327.14    & \textbf{1501.61}   \\ \hline
\end{tabular}
\vspace{-2mm}
\end{table}

\begin{figure}[h]
    \centering
    % \vspace{-4mm}
    \includegraphics[width=\linewidth]{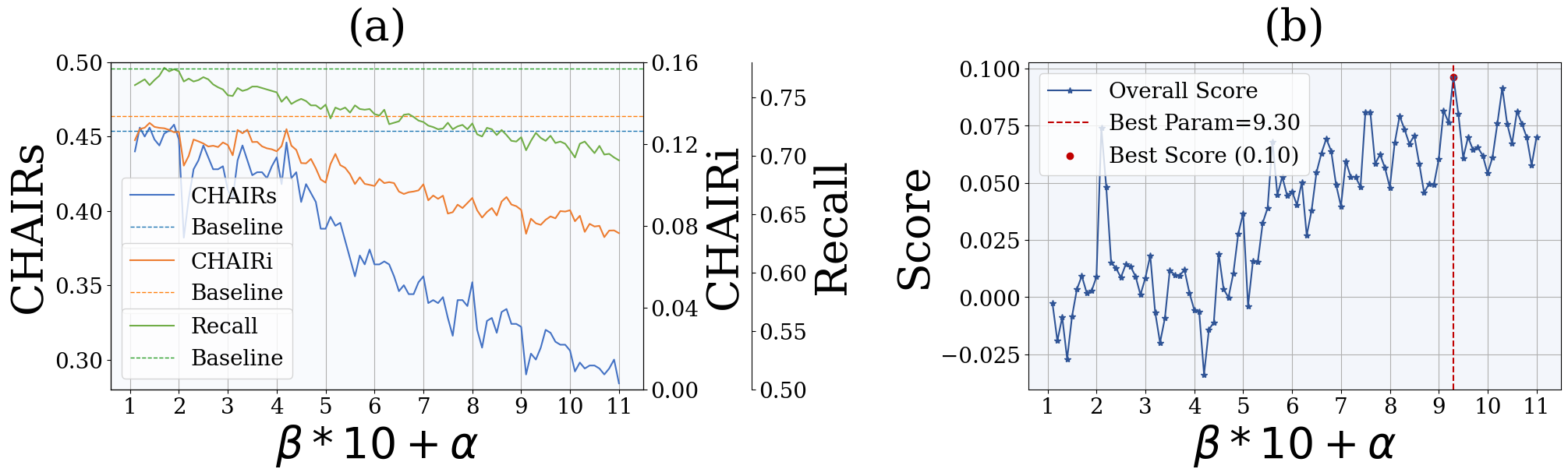}
    \vspace{-8mm}
    \caption{(a) CHAIR scores vs. $\alpha$ and $\beta$ (x-axis: $\beta{\times}10{+}\alpha$). (b) Optimal: $\alpha{=}0.3, \beta{=}0.9$.}
    \label{fig:ablation}
    \vspace{-8mm}
\end{figure}

\noindent\textbf{Hyperparameter Sensitivity Analysis.}
By varying $\alpha$ and $\beta$ in 0.1 intervals with TARAC applied to layers 10-16, we analyze their sensitivity on CHAIR performance. Since \textit{recall} decreases with increasing $\alpha$ and $\beta$, we compute a weighted score to balance hallucination reduction and recall preservation by assigning importance weights: $w_I = w_S = 1, w_R = -2$.
\begin{align}
\textit{Score} &= w_I \cdot (C_I - C_I^{\text{baseline}}) \\
& + w_S \cdot (C_S - C_S^{\text{baseline}}) \\
& + w_R \cdot (\textit{Recall} - \textit{Recall}_{\text{baseline}}),
\end{align}
Figure \ref{fig:ablation} shows the hyperparameter sensitivity analysis. We obtain optimal values $\alpha=0.3$ and $\beta=0.9$,
% achieving $\sim$34.0\%, $\sim$39.6\%,  and $\sim$7.1\% improvements in $C_S$, $C_I$, and \textit{recall} respectively compared to greedy search.
achieving $\sim$34.0\%, $\sim$39.6\% reductions in $C_S$, $C_I$, with a $\sim$7.1\% trade-off in recall.
Although these hyperparameters are determined on CHAIR, the consistent results on AMBER and SHR benchmarks indicate good generalizability across different evaluation metrics.

\section{Conclusion}
In this paper, we propose TARAC, a training-free framework that mitigates hallucinations by counteracting visual attention decay through dynamic attention accumulation. TARAC operates as a highly efficient, plug-and-play module with negligible inference overhead. Extensive experiments demonstrate that it consistently outperforms state-of-the-art baselines across diverse benchmarks. A current limitation is the reliance on fixed hyperparameters. Future work will explore adaptive parameter adjustment and integration with training-based alignment methods.

\bibliographystyle{IEEEbib}
\bibliography{icme2025references}

% \newpage
% \appendix
\clearpage
\appendix
% Appendix Overview
This appendix provides supplementary materials that support the main findings presented in this paper. Section A presents an empirical analysis demonstrating the relationship between hallucinations and attention to image tokens. Section B provides additional evaluation results on AMBER benchmark and detailed layer selection analysis. Section C visualizes the impact of TARAC on the model's attention mechanism through heatmaps and attention distribution curves.

\section*{A Analysis of Hallucination and Visual Attention} 
\label{appendix:analysis}

\begin{figure}[h]
    \centering
    \includegraphics[width=\linewidth]{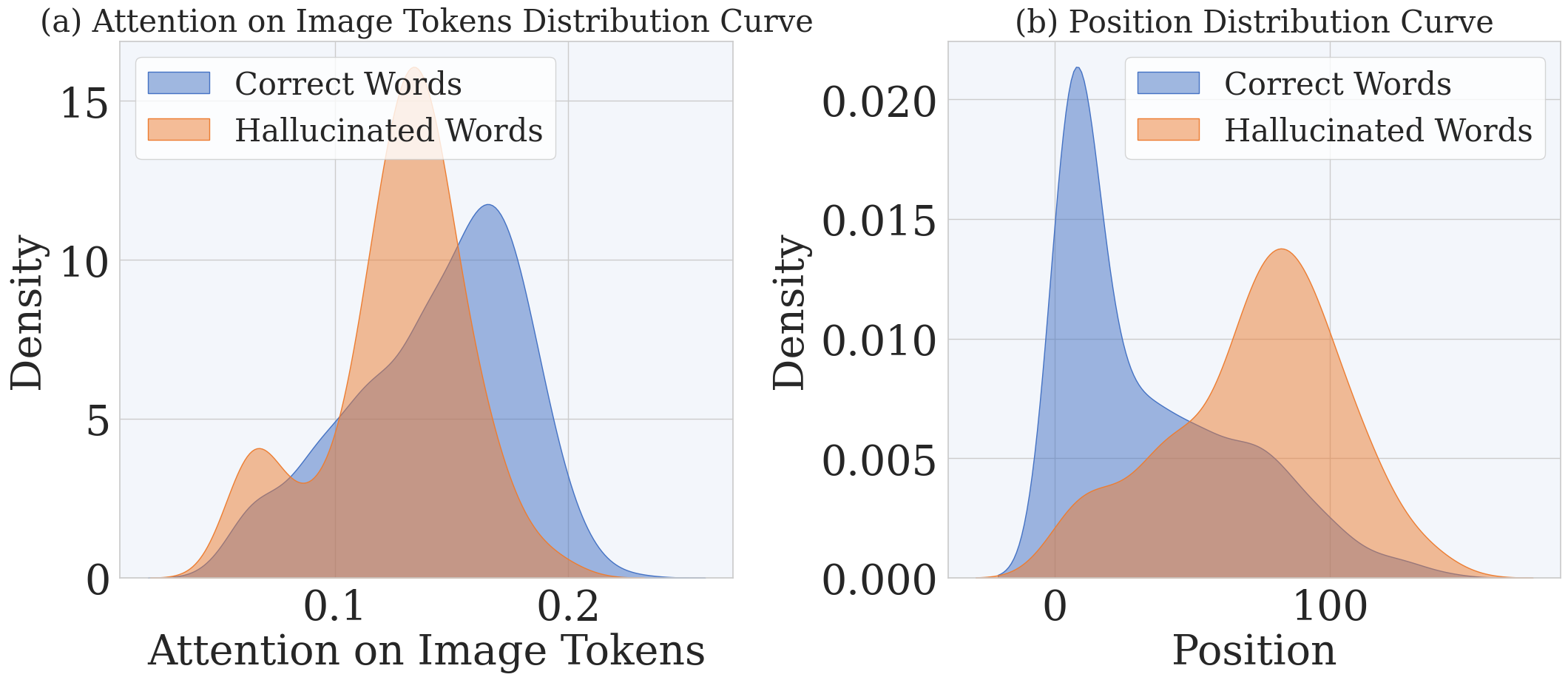}
    \caption{The relationship between hallucinations and attention to image tokens. Only the first occurrence of correct/hallucinated tokens is considered. Figure (a) shows that hallucinated tokens generally exhibit lower attention to image tokens compared to correct tokens. Figure (b) shows that hallucinated tokens are more likely to appear later in the generated captions. Both figures employ Gaussian Kernel Density Estimation, based on 945 correct word samples and 194 hallucinated word samples.}
    \label{fig:distribution}
\end{figure}

To investigate the effects of hallucinations and the decay of attention to image tokens, we use LLaVA-1.5-7B to annotate 500 images from the COCO dataset. During the annotation, we record the attention of current generating token to image tokens for subsequent analysis. We also use the CHAIR evaluation script to identify the words corresponding to hallucinated/correctly annotated objects. Based on the positions of these words in the caption and the previously recorded attention, we compute the visual attention of hallucinated/correct tokens and analyze their distribution. 
As shown in Figure \ref{fig:distribution}, we can clearly observe a negative correlation between visual attention when generating tokens and the probability of the token being a hallucination.

In our analysis, to calculate visual attention, we aggregate the attention scores across all image tokens and compute the average across different layers and attention heads. For both hallucinated and correct tokens, we exclusively consider their first occurrence within the caption and the corresponding visual attention. This is because subsequent repetitions of tokens may introduce bias from the model's language generation capabilities, as these repetitions can often be inferred from the initial occurrence. Additionally, we exclude words that are split into multiple tokens by the tokenizer to avoid introducing noise from the model’s language reasoning priors, which could distort the analysis.

\begin{figure}[h]
    \centering
    \includegraphics[width=\linewidth]{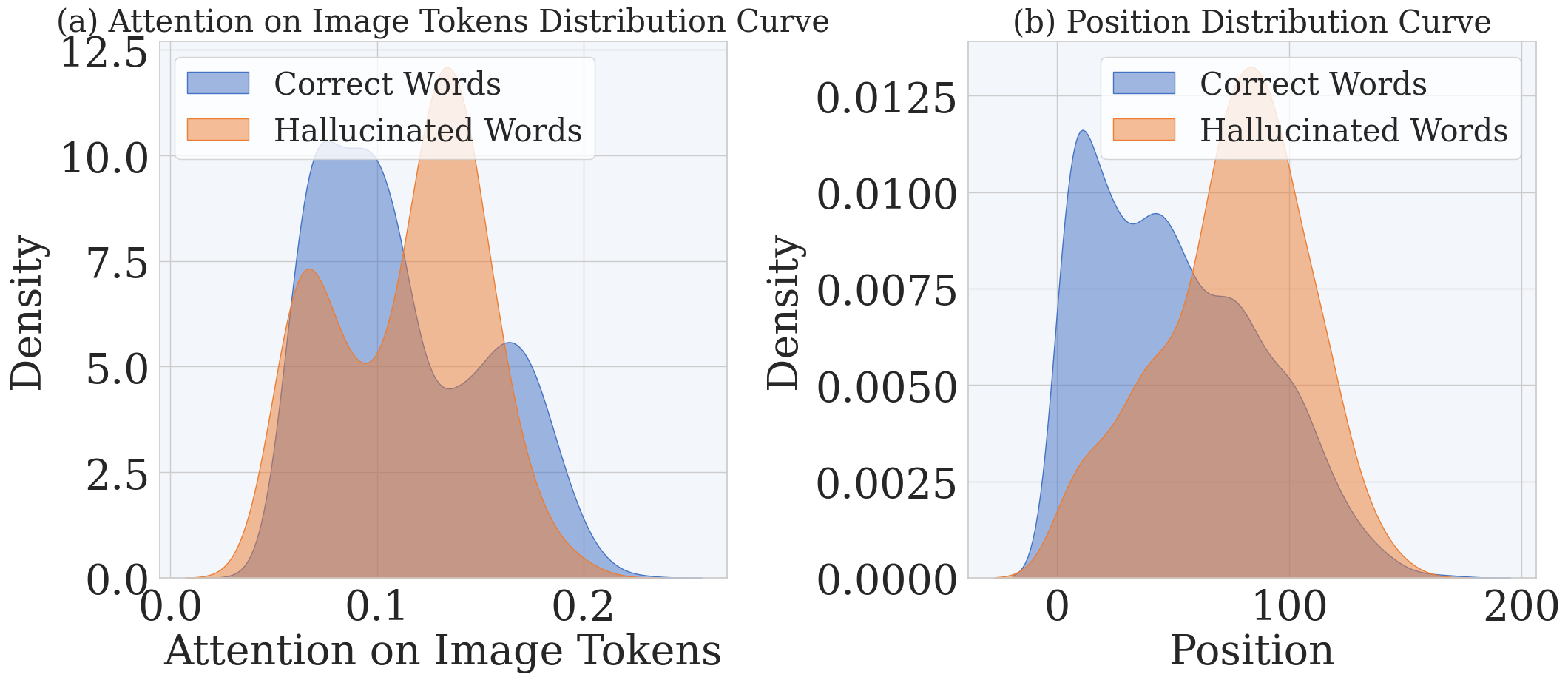}
    \caption{The relationship between hallucinations and attention to image tokens. All occurrences of correct and hallucinated tokens are considered. Figure (a) shows that after including repeated hallucinated and correct tokens, both exhibit higher distribution probabilities in regions with lower attention weights to image tokens. Figure (b) illustrates that the positional distribution of both token types becomes more uniform. Both figures utilize Gaussian kernel density estimation, based on 2,119 correct word samples and 256 hallucinated word samples.}
    \label{fig:distribution-all}
\end{figure}

This point can be inferred by combining Figure \ref{fig:distribution} and Figure \ref{fig:distribution-all}. In Figure \ref{fig:distribution-all}, we observe that after introducing repeated correct tokens, the correct tokens exhibit a distribution peak in the low-attention region, which was not seen in Figure \ref{fig:distribution}. This phenomenon is influenced by the model’s language generation ability, where repeated tokens, despite having low attention on image tokens, can be inferred from the preceding correct token. As a result, the correct tokens have a higher distribution probability in the low-attention area. The same applies to hallucinated tokens. Due to the difference in sample sizes between the two types of tokens, correct tokens form a higher distribution probability in the low-value region. Additionally, the introduction of repeated tokens also leads to a more uniform distribution of both types of tokens across positions.

\section*{B. Additional Evaluation Results}
\label{appendix:amber}
This section provides additional experimental results that complement the main text, including evaluations on AMBER and detailed layer selection analysis.

\subsection*{AMBER Evaluation across Models} 
In addition to the CHAIR results presented in the main text, we also evaluate the generalizability of TARAC across different models using the AMBER benchmark. As shown in Table \ref{tab:amber-model}, TARAC consistently reduces the Hallucination Rate (Hal) and Cognition Score (Cog) across LLaVA-1.5, Qwen2-VL, and InternVL2, demonstrating its robustness.

\begin{table}[h]
\centering
\caption{AMBER on different models w/ and w/o TARAC.}
\label{tab:amber-model}
\vspace{-2mm}
\begin{tabular}{p{1cm}P{0.8cm}P{1.2cm}P{1.2cm}P{0.8cm}P{0.8cm}}
\hline
\footnotesize Model     & \small Method &  \textbf{\scriptsize $\mathrm{CHAIR}(\%)\downarrow$} & \textbf{\scriptsize $\mathrm{Cover}(\%)\uparrow$} & \textbf{\scriptsize $\mathrm{Hal}(\%)\downarrow$}  & \textbf{\scriptsize $\mathrm{Cog}(\%)\downarrow$} \\ \hline
\multirow{2}{*}{\footnotesize LLaVA-1.5} & \footnotesize Greedy & 7.6   & \textbf{49.5}  & 32.1 & 3.8 \\
          & \small Ours  & \textbf{5.0}   & 48.3  & \textbf{27.1} & \textbf{2.5} \\ \hline
\multirow{2}{*}{\footnotesize Qwen2VL}   & \footnotesize Greedy & 5.2   & \textbf{67.0}  & 41.4 & 3.7 \\
          & \small Ours  & \textbf{4.1}   & 53.3  & \textbf{21.1} & \textbf{2.0} \\ \hline
\multirow{2}{*}{\footnotesize InternVL2} & \footnotesize Greedy & 8.5   & \textbf{73.4}  & 69.1 & 8.8 \\
          & \small Ours  &   \textbf{8.5}  & 70.4  & \textbf{63.6} & \textbf{8.2} \\ \hline
\end{tabular}
\vspace{-2mm}
\end{table}

\begin{table*}[!h]
\centering
\caption{Detailed MME performance on LLaVA-1.5 with TARAC applied to different layer ranges.}
\label{tab:mme-layer-full}
\vspace{-2mm}
\renewcommand{\arraystretch}{1.2}
\begin{tabular}{ccccccccc}
\hline
\multirow{3}{*}{\small Layer Range}     & \multirow{3}{*}{\small Cognition$\uparrow$} & \multicolumn{7}{c}{Perception$\uparrow$}                                                                                                                                                                    \\ \cline{3-9} 
                           &                            & \multirow{2}{*}{OCR} & \multirow{2}{*}{\parbox{1.1cm}{\footnotesize Fine-\vspace{-0.7ex} \\Grained\vspace{-0.7ex} \\ Subset}} & \multicolumn{5}{c}{Hallucination Subset}                                                 \\ \cline{5-9} 
                           &                            &                      &                                                                                 & existence       & count           & position        & color           & total            \\ \hline
baseline                & 323.57                     & \textbf{132.50}               & 720.47                               & 190.00    & 143.33 & \textbf{133.33}   & 163.33 & 1482.96 \\
{[}1:10{]}              & 307.86                     & 122.50               & \underline{724.34}                               & 190.00    & 143.33 & 123.33   & 163.33 & 1466.83 \\
    {[}8:16{]}              & \underline{332.86}                     & \textbf{132.50}               & 720.73                               & 190.00    & \textbf{163.33} & 123.33   & \textbf{170.00} & \underline{1499.89} \\
{[}10:16{]}             & \textbf{339.29}                     & \textbf{132.50}               & 722.98                               & 190.00    & \textbf{163.33} & 123.33   & \underline{165.00} & 1497.14 \\
{[}10:18{]}             & 327.14                     & \underline{125.00}               & 722.61                               & 190.00    & \textbf{163.33} & 123.33   & \textbf{170.00} & 1494.27 \\
{[}10:20{]}             & 327.14                     & \underline{125.00}               & \textbf{724.95}                               & \textbf{195.00}    & \textbf{163.33} & 123.33   & \textbf{170.00} & \textbf{1501.61} \\
\hline
\end{tabular}
\vspace{-2mm}
\end{table*}

\subsection*{Detailed Layer Selection Analysis} \label{appendix:layer}
Table \ref{tab:mme-layer-full} presents the detailed breakdown of MME scores for different layer ranges, complementing the summarized results in the main text.

\section*{C. Impact of TARAC on Visual Attention}
\label{appendix:vis}
\begin{figure}[h]
    \centering
    \includegraphics[width=0.92\linewidth]{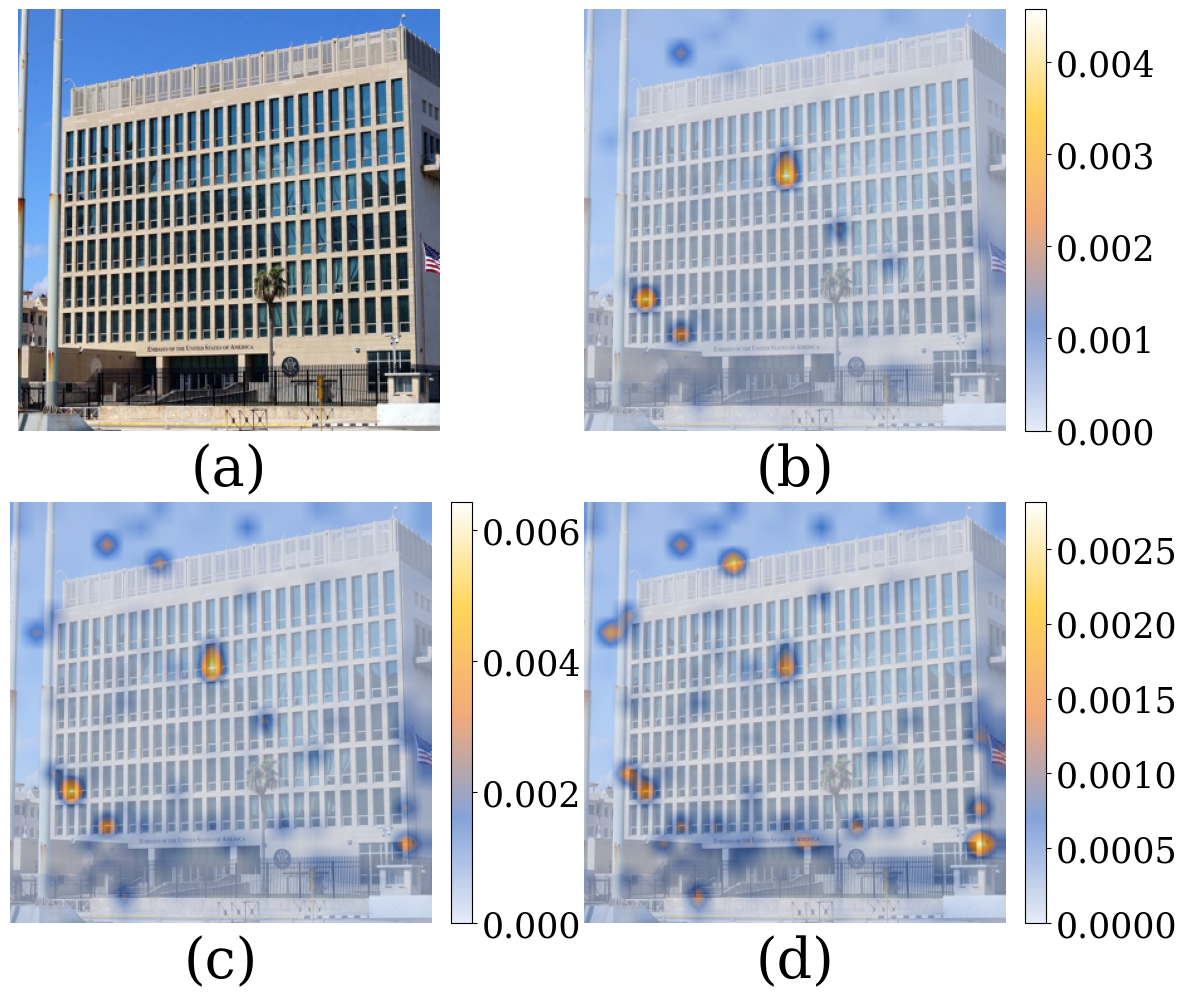}
    \caption{\textbf{Impact of TARAC on visual attention.} Attention averaged across layers, tokens, and heads. (a) Original image, (b) LLaVA baseline, (c) LLaVA with TARAC, (d) Attention difference.}
    \label{fig:heatmap}
\end{figure}

\begin{figure}[h]
    \centering
    \includegraphics[width=0.95\linewidth]{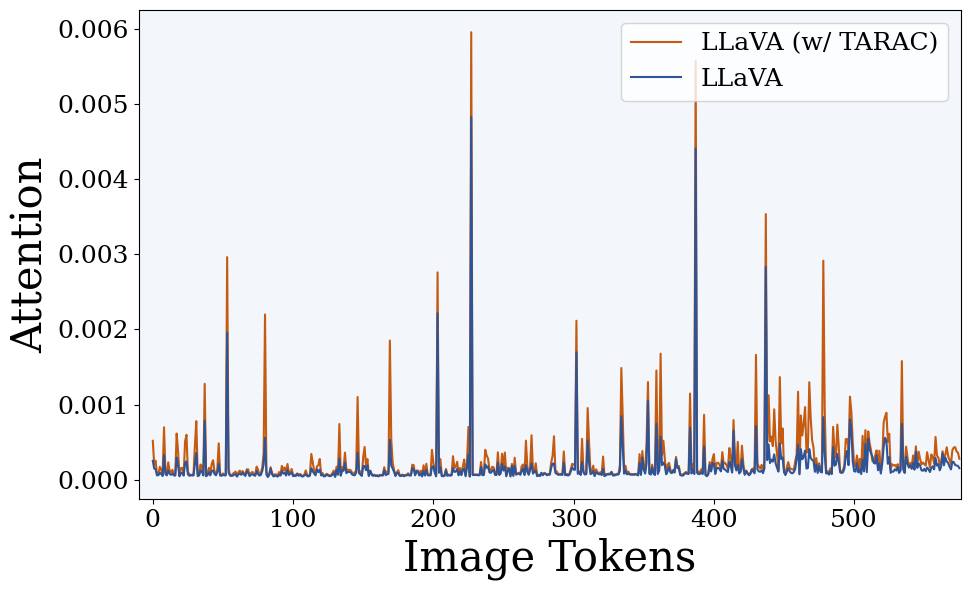}
    \caption{\textbf{Attention distribution curves w/ and w/o TARAC.} Attention averaged across layers, tokens, and heads.}
    \label{fig:curve}
\end{figure}

To visualize the effect of TARAC on the model's attention mechanism, we present attention heatmaps and attention distribution curves. Figure \ref{fig:heatmap} shows the spatial attention patterns before and after applying TARAC, while Figure \ref{fig:curve} illustrates the quantitative changes in attention weights across image tokens.

% \vspace{12pt}
% \color{red}
% IEEE conference templates contain guidance text for composing and formatting conference papers. Please ensure that all template text is removed from your conference paper prior to submission to the conference. Failure to remove the template text from your paper may result in your paper not being published.

\end{document}